\documentclass{article} 

\usepackage[table]{xcolor}
\definecolor{skyColor}{rgb}{0.4157,0.5333,0.8000}
\definecolor{buildingColor}{rgb}{0.9137,0.3490,0.1882}
\definecolor{poleColor}{rgb}{1, 0.5, 0}
\definecolor{roadColor}{rgb}{1.0000,0.8078,0.8078}    
\definecolor{sidewalkColor}{rgb}{0.9412,0.1373,0.9216} 
\definecolor{vegetationColor}{rgb}{0, 0.8549, 0}     
\definecolor{tsignColor}{rgb}{0.4588,0.1137,0.1608} 
\definecolor{fenceColor}{rgb}{0.4157,0.5333,0.0}
\definecolor{carColor}{rgb}{0.5843,0,0.9412}       
\definecolor{pedestrianColor}{rgb}{0.8706,0.9451,0.0941} 
\definecolor{cyclistColor}{rgb}{0,0.6549,0.6118}
\definecolor{voidColor}{rgb}{1,1,1}

\usepackage{iclr2016_workshop,times}

\usepackage{url}
\usepackage{graphicx}
\usepackage{caption}
\usepackage{amsmath}
\usepackage{amssymb}
\usepackage{hyperref}

\title{Spatio-Temporal Video Autoencoder with \\ Differentiable Memory}

\author{Viorica P{\u a}tr{\u a}ucean, Ankur Handa \& Roberto Cipolla \\
Department of Engineering\\
University of Cambridge, UK\\
\texttt{\{vp344,ah781,rc10001\}@cam.ac.uk} \\
}

\newcommand{\eg}{\emph{e.g.~}}
\newcommand{\ie}{\emph{i.e.~}}

\newcommand{\wrt}{w.r.t.~}

\begin{document}

\maketitle

\begin{abstract}
We describe a new spatio-temporal video autoencoder, based on a classic spatial image autoencoder and a novel nested temporal autoencoder. The temporal encoder is represented by a differentiable visual memory composed of convolutional long short-term memory (LSTM) cells that integrate changes over time. Here we target motion changes and use as temporal decoder a robust optical flow prediction module together with an image sampler serving as built-in feedback loop. The architecture is end-to-end differentiable. At each time step, the system receives as input a video frame, predicts the optical flow based on the current observation and the LSTM memory state as a dense transformation map, and applies it to the current frame to generate the next frame. By minimising the reconstruction error between the predicted next frame and the corresponding ground truth next frame, we train the whole system to extract features useful for motion estimation without any supervision effort. We present one direct application of the proposed framework in weakly-supervised semantic segmentation of videos through label propagation using optical flow. 

\end{abstract}

\section{Introduction}
\label{sec:intro}
High-level understanding of video sequences is crucial for any autonomous intelligent agent, \eg semantic segmentation is indispensable for navigation, or object detection skills condition any form of interaction with objects in a scene. The recent success of convolutional neural networks in tackling these high-level tasks for static images opens up the path for numerous applications. However, transferring the capabilities of these systems to tasks involving video sequences is not trivial, on the one hand due to the lack of video labelled data, and on the other hand due to convnets' inability of exploiting temporal redundancy present in videos. Motivated by these two shortcomings, we focus on reducing the supervision effort required to train recurrent neural networks, which are known for their ability to handle sequential input data \cite{Williams1995,HochreiterEtAl2000}.    

In particular, we describe a spatio-temporal video autoencoder integrating a differentiable short-term memory module whose (unsupervised) training is geared towards motion estimation and prediction \cite{Horn:1994}. This choice has biological inspiration. The human brain has a complex system of visual memory modules, including iconic memory, visual short-term memory (VSTM), and long-term memory \cite{Hollingworth04}. Among them, VSTM is responsible mainly for understanding visual changes (movement, light changes) in dynamic environments, by integrating visual stimuli over periods of time \cite{Phillips74,Magnussen2000247}. Moreover, the fact that infants are able to handle occlusion, containment, and covering events by the age of 2.5 months \cite{baillargeon04} could suggest that the primary skills acquired by VSTM are related to extracting features useful for motion understanding. These features in turn could generate objectness awareness based on the simple principle that points moving together belong to the same object. Understanding objectness is crucial for high-level tasks such as semantic segmentation or action recognition \cite{Alexe2012}. In this spirit, our approach is similar to the recent work of \cite{AgrawalCM15}, who show that the features learnt by exploiting (freely-available) ego-motion information as supervision data are as good as features extracted with human-labelled supervision data. We believe that our work is complementary to their approach and integrating them could lead to  an artificial agent with enhanced vision capabilities.    

Our implementation draws inspiration from standard video encoders and compression schemes, and suggests that deep video autoencoders should differ conceptually from spatial image autoencoders. A video autoencoder need not reproduce \emph{by heart} an entire video sequence. Instead, it should be able to encode the significant differences that would allow it to reconstruct a frame given a previous frame. To this end, we use a classic convolutional image encoder -- decoder with a nested memory module composed of convolutional LSTM cells, acting as temporal encoder. Since we focus on learning features for motion prediction, we use as temporal decoder a robust optical flow prediction module together with an image sampler, which provides immediate feedback on the predicted flow map. At each time step, the system receives as input a video frame, predicts the optical flow based on the current frame and the LSTM content as a dense transformation map, and applies it to the current frame to predict the next frame. By minimising the reconstruction error between the predicted next frame and the ground truth next frame, we are able to train the whole system for motion prediction without any supervision effort. Other modules handling other types of variations like light changes could be added in parallel, inspired by neuroscience findings which suggest that VSTM is composed of a series of different modules specialised in handling the low-level processes triggered by visual changes, all of them connected to a shared memory module \cite{Magnussen2000247}. Note that at the hardware level, this variations-centred reasoning is similar to event-based cameras \cite{Boahen05}, which have started to make an impact in robotic applications \cite{kim2014}.           

Summary of contributions: We propose a spatio-temporal version of LSTM cells to serve as a basic form of visual short-term memory, and make available an end-to-end differentiable architecture with built-in feedback loop, that allows effortless training and experimentation with the goal of understanding the role of such an artificial visual short-term memory in low-level visual processes and its implications in high-level visual tasks. We present one direct application of the proposed framework in weakly-supervised semantic segmentation of videos through label propagation using optical flow. All our code (Torch implementation) is available online \cite{ConvLSTM_code}.  

\section{Related Work}
\label{sec:relatedwork}
Architectures based on LSTM cells \cite{Hochreiter1997} have been very successful in various tasks involving one-dimensional temporal sequences: speech recognition \cite{SakSB14}, machine translation \cite{SutskeverVL14}, music composition \cite{MSU-CSE-00-2}, due to their ability to preserve information over long periods of time. Multi-dimensional LSTM networks have been proposed to deal with (2D) images \cite{Graves2007} or (3D) volumetric data \cite{StollengaBLS15}, treating the data as spatial sequences. Since in our work we aim at building a visual short-term memory, customised LSTM cells that deal with temporal sequences of spatial data represent a natural choice.   

Recently, \cite{SrivastavaMS15} proposed an LSTM-based video autoencoder, which aims at generating past and future frames in a sequence, in an unsupervised manner. However, their LSTM implementation treats the input as sequences of vectors by flattening the frames or by using 1D frame representations produced after the last fully-connected layer of a convolutional neural network. This results in a large number of parameters, since the activations of the LSTM cells use fully-connected linear operations, which are not necessarily useful for 2D images, since natural image statistics indicate only local correlations. Another difference is in the architecture setup. We are not interested in training a black-box to produce past and future frames in the sequence. Instead we aim at a transparent setup, and train a more generic memory module together with specialised modules able to decode the memory content and on which appropriate constraints (sparsity, smoothness) can be imposed.  

Our approach is partially related to optical flow estimation works like DeepFlow \cite{Weinzaepfel13} and FlowNet \cite{FischerDIHHGSCB15}. However, these methods use purely supervised training to establish matches between consecutive pairs of frames in a sequence, and then apply a variational smoothing. Our architecture is end-to-end differentiable and integrates a smoothness penalty to ensure that nearby pixels follow a locally smooth motion, and requires no labelled training data.     

As mentioned in the previous section, our work is similar in spirit to \cite{AgrawalCM15}, by establishing a direct link between vision and motion, in an attempt to reduce supervision effort for high-level scene understanding tasks. However, instead of relying on data provided by an inertial measurement unit from which we can estimate only a global transformation representing the ego-motion, we predict a dense flow map by integrating visual stimuli over time using a memory module, and then we use a built-in feedback loop to assess the prediction. The dense flow map is useful to explain dynamic environments, where different objects can follow different motion models, hence a global transformation is not sufficient \cite{menze2015object}.

\section{Architecture}
\label{sec:architecture}
\begin{figure}[t]
\begin{center}
    \centering
    \includegraphics[width=\textwidth]{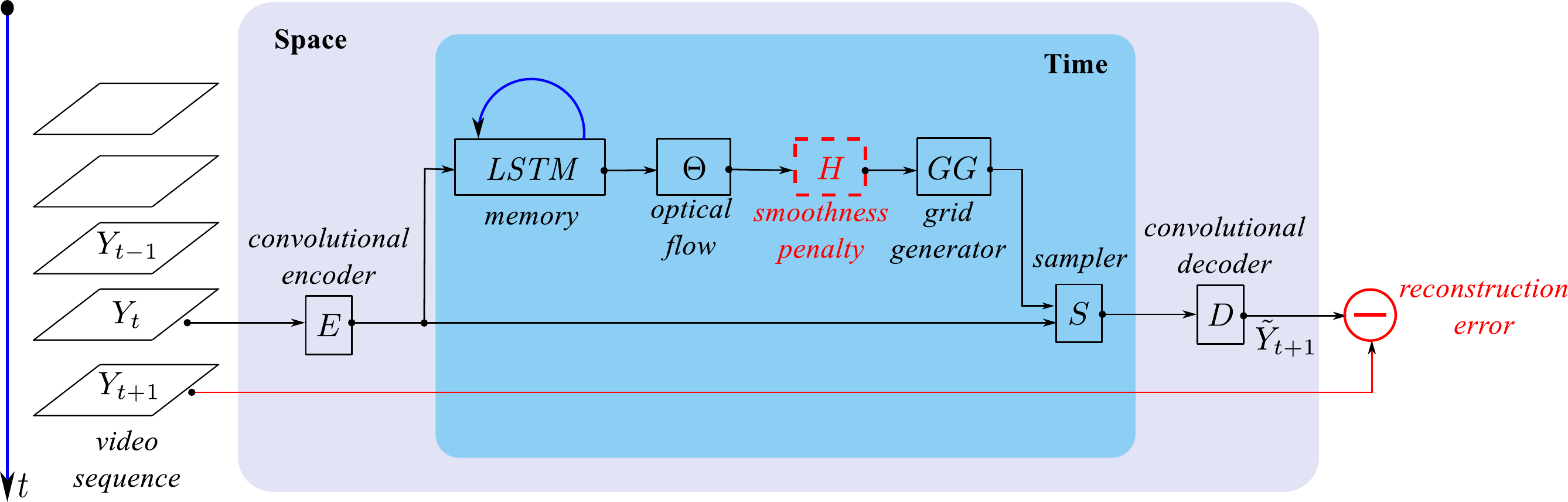}
    \caption{Spatio-temporal video autoencoder.}
    \label{fig:tae}
\end{center}%
\end{figure}
Our architecture consists of a temporal autoencoder nested into a spatial autoencoder (see Figure \ref{fig:tae}). At each time step, the network takes as input a video frame $Y_t$ of size $H\times W$, and generates an output of the same size, representing the predicted next frame, $\tilde{Y}_{t+1}$. In the following, we describe each of the modules in detail. 

\subsection{Spatial autoencoder $E$ and $D$}
The spatial autoencoder is a classic convolutional encoder -- decoder architecture.
The encoder $E$ contains one convolutional layer, followed by $\tanh$ non-linearity and a spatial max-pooling with subsampling layer. The decoder $D$ mirrors the encoder, except for the non-linearity layer, and uses nearest-neighbour spatial upsampling to bring the output back to the size of the original input. After the forward pass through the spatial encoder $ Y_t \xrightarrow{E} x_t  $, the size of the feature maps $x_t$ is $d\times h\times w$, $d$  being the number of features, and $h$ and $w$ the height and width after downsampling, respectively. 

\subsection{Temporal autoencoder}
The goal of the temporal autoencoder is to capture significant changes due to motion (ego-motion or movement of the objects in the scene), that would allow it to predict the visual future, knowing the past and the present. In a classic spatial autoencoder \cite{Masci2011}, the encoder and decoder learn proprietary feature spaces that allow an optimal decomposition of the input using some form of regularisation to prevent learning a trivial mapping. The encoder decides freely upon a decomposition based on its current feature space, and the decoder constrains the learning of its own feature space to satisfy this decomposition and to reconstruct the input, using usually operations very similar to the encoder, and having the same number of degrees of freedom. Differently from this, the proposed temporal autoencoder has a decoder with a small number of trainable parameters, whose role is mainly to provide immediate feedback to the encoder, but without the capacity of amending encoder's mistakes like in the spatial case. In optimisation terms, the error during learning is attributed mainly to the encoder, which is now more constrained to produce sensible feature maps.

\subsubsection{Memory module $\mathit{LSTM}$}
The core of the proposed architecture is the memory module playing the role of a temporal encoder. We aim at building a basic visual short-term memory, which preserves locality and layout ensuring a fast access and bypassing more complicated addressing mechanisms like those used by Neural Turing Machines \cite{GravesWD14}. To this end, we use customised spatio-temporal LSTM cells with the same layout as the input. At each time step $t$, the $\mathit{LSTM}$ module receives as input a new video frame after projection in the spatial feature space. This is used together with the memory content and output of the previous step $t-1$ to compute the new memory activations.
%

Classic LSTM cells operate over sequences of (one-dimensional) vectors and perform biased linear (fully-connected) transformations, followed by non-linearities to compute gate and cell activations. In our case, to deal with the spatial and local nature of the video frames, we replace the fully-connected transformations with spatial local convolutions. 
Therefore, the activations of a spatio-temporal convolutional LSTM cell at time $t$ are given by:

\begin{equation}
\left\{ \begin{array}{ccl}
i_t & = & \sigma(x_t \ast w_{xi} + h_{t-1} \ast w_{hi} + w_{i\text{bias}}) \\
f_t & = & \sigma(x_t \ast w_{xf} + h_{t-1} \ast w_{hf} + w_{f\text{bias}}) \\
\tilde{c}_t & = & \tanh(x_t \ast w_{x\tilde{c}} + h_{t-1} \ast w_{h\tilde{c}} + w_{\tilde{c}\text{bias}}) \\
c_t & = & \tilde{c}_t \odot i_t + c_{t-1} \odot f_t \\
o_t & = & \sigma(x_t \ast w_{xo} + h_{t-1} \ast w_{ho} + w_{o\text{bias}}) \\
h_t & = & o_t \odot \tanh(c_t) \\ 
\end{array} \right.
\label{eq:lstm}
\end{equation}


where $x_t$ represents the input at time $t$, \ie the feature maps of the frame $t$; $i_t$, $f_t$, $\tilde{c}_t$, and $o_t$ represent the input, forget, cell, and output gates, respectively; $c_t$, $c_{t-1}$, $h_t$, and $h_{t-1}$ are the memory and output activations at time $t$ and $t-1$, respectively; $\sigma$ and $\tanh$ are the sigmoid and hyperbolic tangent non-linearities; $\ast$ represents the convolution operation, and $\odot$ the Hadamard product. For input feature maps of size $d\times h \times w$, the $\mathit{LSTM}$ module outputs a memory map of size $d_m \times h \times w$, where $d_m$ is the number of temporal features learnt by the memory. The recurrent connections operate only over the temporal dimension, and use local convolutions to capture spatial context, unlike multi-dimensional LSTM versions that use spatial recurrent connections \cite{Graves2007,StollengaBLS15}. 
Note that a similar convolutional LSTM implementation was recently used in \cite{ShiCWYWW15} for precipitation nowcasting.

\subsubsection{Optical flow prediction $\Theta$ with Huber penalty $H$}
The optical flow prediction module generates a dense transformation map $\mathcal{T}$, having the same height and width as the memory output, with one 2D flow vector per pixel, representing the displacement in $x$ and $y$ directions due to motion between consecutive frames. $\mathcal{T}$ allows predicting the next frame by warping the current frame. We use two convolutional layers with relatively large kernels ($15\times 15$) to regress from the memory feature space to the space of flow vectors. Large kernels are needed since the magnitude of the predicted optical flow is limited by the size of the filters. To ensure local smoothness, we need to penalise the local gradient of the flow map $\triangledown \mathcal{T}$. We add a penalty module $H$ whose role is to forward its input unchanged during the forward pass, and to inject non-smoothness error gradient during the backward pass, towards the modules that precede it in the architecture and that might have contributed to the error. We use Huber loss as penalty, with its corresponding derivative (\ref{eq:huber}), due to its edge-preserving capability \cite{Werlberger2009a}. The gradient of the flow map is obtained by convolving the map with a fixed (non-trainable) $2\times3\times3$ filter, corresponding to a 5-point stencil, and $0$ bias.  
\begin{equation}
\mathcal{H}_{\delta}(a_{ij}) = \left\{ \begin{array}{ll} \frac{1}{2} a_{ij}^2, & \text{for } |a_{ij}| \le \delta \\
                                          \delta(|a_{ij}| - \frac{1}{2}\delta), & \text{otherwise} 
                                          \end{array}, \right. 
                                          \nabla \mathcal{H}_{\delta}(a_{ij}) = \left\{ \begin{array}{ll} a_{ij} & \text{for} |a_{ij}| \le \delta, \\
                                          \delta \text{sign}(a_{ij}), & \text{otherwise} 
                                          \end{array} \right.
\label{eq:huber}
\end{equation}
Here, $a_{ij}$ represent the elements of $\triangledown \mathcal{T}$. In our experiments, we used $\delta=10^{-3}$.   

\subsubsection{Grid generator $\mathit{GG}$ and image sampler $S$}
The grid generator $\mathit{GG}$ and the image sampler $S$ output the predicted feature maps $\tilde{x}_{t+1}$ of the next frame after warping the current feature maps $x_t$ with the flow map produced by the $\Theta$ module. We use similar differentiable grid generator and image sampler as Spatial Transformer Network (STN) \cite{JaderbergSZK15} (implementation available online \citep{STNImplementation}). The output of $S$, of size $d \times h \times w$, is considered as a fixed $h\times w$ grid, holding at each $(x_o,y_o)$ position a feature map entry of size $1 \times 1 \times d$. We modified the grid generator to accept one transformation per pixel, instead of a single transformation for the entire image as in STN. Given the flow map $\mathcal{T}$, $\mathit{GG}$ computes for each element in the grid the source position $(x_s,y_s)$ in the input feature map from where $S$ needs to sample to fill in the position $(x_o, y_o)$:
\begin{equation}
\left( \begin{array}{c}
x_s \\ y_s
 \end{array} \right) = \mathcal{T} (x_o,y_o) \left( \begin{array}{c}
x_o \\ y_o \\ 1
 \end{array} \right), \mathcal{T}(\cdot,\cdot) = \left( \begin{array}{ccc}
1 & 0 & t_x \\ 0 & 1 & t_y
\end{array} \right). 
\label{eq:gg}
\end{equation}


The $\Theta$ module outputs two parameters $(t_x,t_y)$ for each pixel. The forward pass of $\mathit{GG}$ is given by equation (\ref{eq:gg}). In the backward pass, $\mathit{GG}$ simply backpropagates the derivative of equation (\ref{eq:gg}) \wrt the input parameters $(t_x,t_y)$.   

\subsubsection{Loss function}
Training the network comes down to minimising the reconstruction error between the predicted next frame and the ground truth next frame, with Huber penalty gradient injected during backpropagation on the optical flow map:
\begin{equation}
\mathcal{L}_t = \| \tilde{Y}_{t+1} - Y_{t+1} \|_2^2 + w_H\mathcal{H}(\triangledown \mathcal{T}),
\label{eq:loss}
\end{equation}
where $w_H$ is a hard-coded parameter used to weight the smoothness constraints \wrt the data term; in our experiments $w_H = 10^{-2}$, which is a common value in optical flow works.

\subsection{Network parameters}
The proposed network has 1,035,067 trainable parameters.  The spatial encoder and decoder have 16 filters each, size $7\times 7$. The memory module ($\mathit{LSTM}$) has 64 filters, size $7\times 7$, and the optical flow regressor $\Theta$ has 2 convolutional layers, each with 2 filters of size $15\times 15$, and a $1\times 1$ convolutional layer. The other modules: $\mathit{GG}$, $H$, and $S$ have no trainable parameters.


\section{Training}
\label{sec:training}

The training was done using \emph{rmsprop}, with a learning rate starting at $10^{-4}$ and decaying by $0.9$ after every 5 epochs. The initialisation of the convolutional layers except those involved in the memory module was done using \emph{xavier} method \cite{Glorot10}. The parameters of the memory module, except the biases, were initialised from a uniform distribution $\mathcal{U}(-0.08,0.08)$. The biases of the forget gates are initialised to 1; for the other gates we set the biases to 0. During training, before each parameter update, we clip the gradients to lie in a predefined range, to avoid numerical issues \cite{Graves13}.

Our implementation is based on Torch library \cite{Collobert_torch7:a} and extends the \emph{rnn} package \cite{LeonardWW15}. All our code is available online \cite{ConvLSTM_code}. The training was done on an NVIDIA K40 GPU (12G memory). 

\section{Experiments}
\label{sec:experiments}
\begin{figure}[t]
\begin{center}
   \includegraphics[width=.8\linewidth]{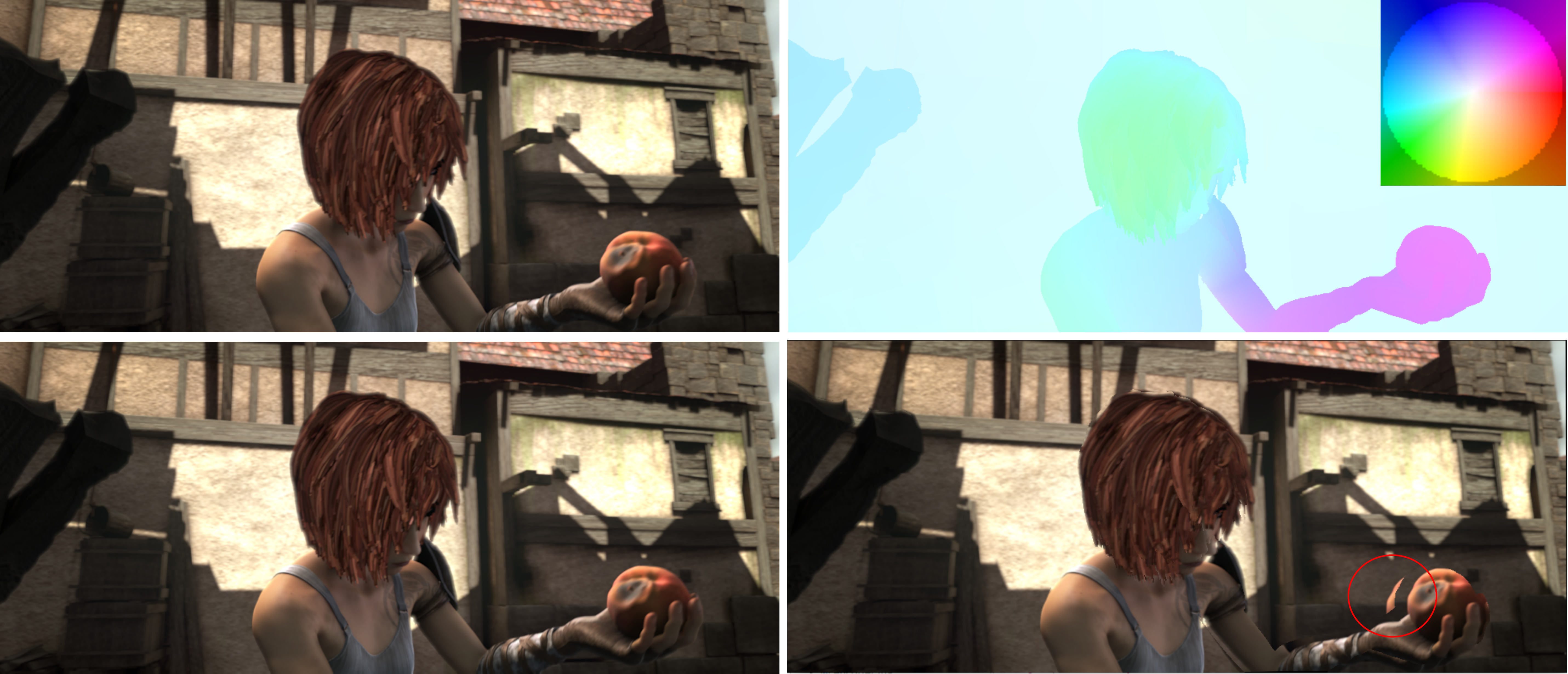}
\end{center}
\caption{Warping performed by the $\mathit{GG}$ and $S$ modules, given a video frame and the ground truth optical flow map. First line: input frame and ground truth optical flow map displayed using the colour code shown in the top-right corner of the image; the colour encodes the direction of movement and the colour intensity reflects the magnitude; darker shades correspond to higher magnitudes \cite{Baker:2011}. Second line: ground truth next frame and sampled next frame; note the artifacts close to the boundaries.}
\label{fig:warp}
\end{figure}

As a sanity check to confirm the ability of the grid generator $\mathit{GG}$ and image sampler $S$ to generate the next frame given the current frame and the correct optical flow map, we ran simple warping tests using Sintel dataset \cite{Butler2012}, by isolating the two modules from the rest of the architecture. Figure \ref{fig:warp} shows an example of warping. Note that since the flow displacement is significant in this dataset, the sampling result can contain artefacts, especially near boundaries, caused by occlusions. The average per-pixel error induced by the sampler on this dataset is $0.004$ (for pixel values in $[0,1]$).  However, this test was done by sampling directly from the input frames, no spatial encoder -- decoder was used. In our architecture, these artefacts will be washed out to a certain extent by the convolutional decoder. The optical flow throughout this section is displayed using the colour code from \cite{Baker:2011}, illustrated in Figure \ref{fig:warp}, top-right corner.

\subsection{Unsupervised experiments}
To assess qualitatively and quantitatively the behaviour of the proposed architecture and of its components, we ran unsupervised experiments on synthetic and real datasets\footnote{Existing datasets in the optical flow community do not represent suitable training and/or testing data because they contain flow maps for only pairs of images \cite{FischerDIHHGSCB15}, or for very few short video sequences (6 sequences of 2 or 8 frames in Middlebury dataset \cite{Baker:2011}). Sintel dataset \cite{Butler2012} contains indeed ground truth optical flow for longer video sequences. However, it exhibits a complex combination of camera self-motion and moving objects, with a relatively low frame rate, leading to very large (non-realistic) displacements between consecutive frames. Note also that the self-motion makes the entire scene to \emph{move} in terms of optical flow, making the prediction very hard and, in the same time, not useful for the purposes of our work: if all the points in the scene move, it is difficult to delineate objects.}.

\begin{figure}[t]
\begin{center}
   \includegraphics[width=\linewidth]{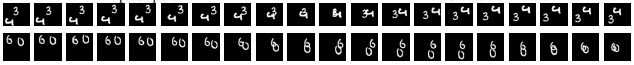}\\
\end{center}
   \caption{Samples of images from moving MNIST dataset.}
\label{fig:samples}
\end{figure} 

\paragraph{Synthetic dataset.}
Moving MNIST dataset \cite{SrivastavaMS15} consists of sequences of 20 frames each, obtained by moving (translating) MNIST digit images inside a square of size $64\times 64$, using uniform random sampling to obtain direction and velocity; the sequences can contain several overlapping digits in one frame. We generated 10k sequences for training and 3k sequences for validation (see Figure \ref{fig:samples} for sample images).  

\begin{figure}[t]
\begin{center}
   \includegraphics[width=\linewidth]{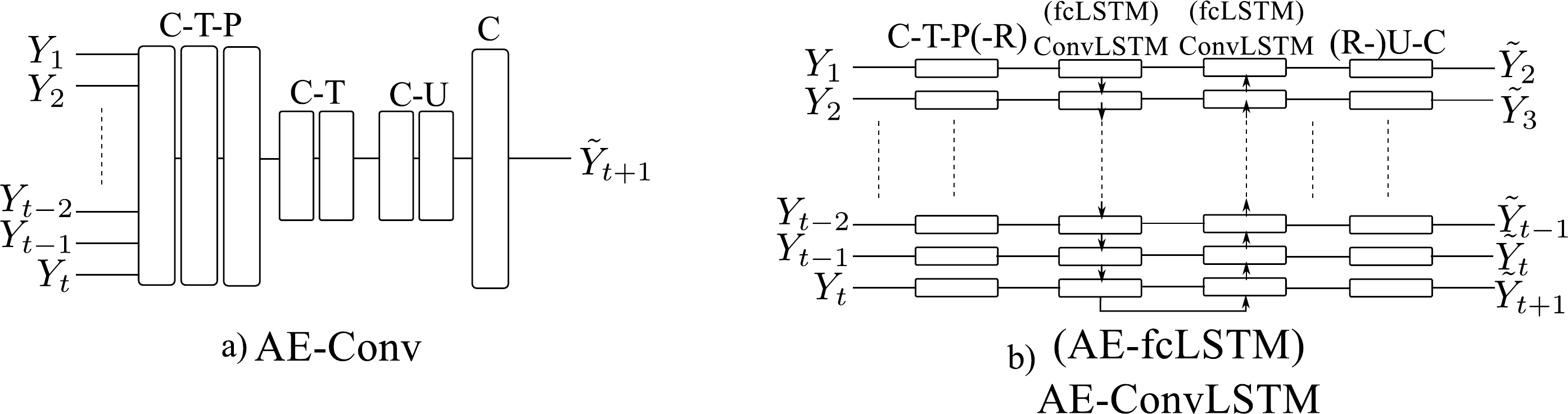}
\end{center}
   \caption{Baseline (autoencoder) next frame predictors. C = convolutional, T = $\tanh$, P = pooling, U = unpooling, R = reshape from 2D feature maps to 1D (flattened) vectors and vice-versa.}
\label{fig:baselinesAE} 
\end{figure}

\begin{table}[t!]
\begin{center}
\begin{tabular}{|l|c|c|}
\hline
\textbf{Architecture} & \textbf{Parameters} & \textbf{Test error} \\
\hline
AE-Conv & 109,073 & 0.0948 \\ 
\hline
AE-fcLSTM & 33,623,649 & 0.0650 \\
\hline
AE-ConvLSTM & 1,256,305 & 0.0641\\ 
\hline
AE-ConvLSTM-flow & 1,035,067 & \textbf{0.0439} \\
\hline
\end{tabular}
\end{center}
\caption{Number of parameters for the different architectures and their per-pixel average error on moving MNIST.}
\label{tab:all_params}
\end{table}

To highlight the advantages of the proposed convolutional LSTM units (denoted as ConvLSTM) and of the overall architecture (denoted as AE-ConvLSTM-flow), we ran quantitative experiments using different architectures, illustrated in Figure \ref{fig:baselinesAE}. AE-Conv is a simple convolutional autoencoder, which takes as input the frames of a video sequence considered as independent channels, and outputs the predicted next frame. It contains a similar encoder and decoder as AE-ConvLSTM-flow, but replaces the (recurrent) memory module and optical flow module with simple convolution-tanh layers. AE-ConvLSTM is a similar setup, but uses convolutional LSTM blocks as temporal encoder and decoder respectively; AE-fcLSTM replaces the convolutional LSTM blocks with classic fully-connected LSTM blocks, resembling the architecture of \cite{SrivastavaMS15}, but predicting only one frame in the future. Table \ref{tab:all_params} summarises the number of parameters for each architecture. Note that, although the depth of all the architectures could be considered similar, the differences in the number of parameters are significant: all the architectures using LSTM blocks have more parameters than the simple convolutional autoencoder due to the gating layers; among these, the fully-connected model is considerably larger. AE-ConvLSTM-flow has less parameters than AE-ConvLSTM due to the fact that the LSTM temporal decoder in the latter is replaced by simple convolutional regression layers in the former. 

\begin{figure}[t]
\begin{center}
   \includegraphics[width=.8\linewidth]{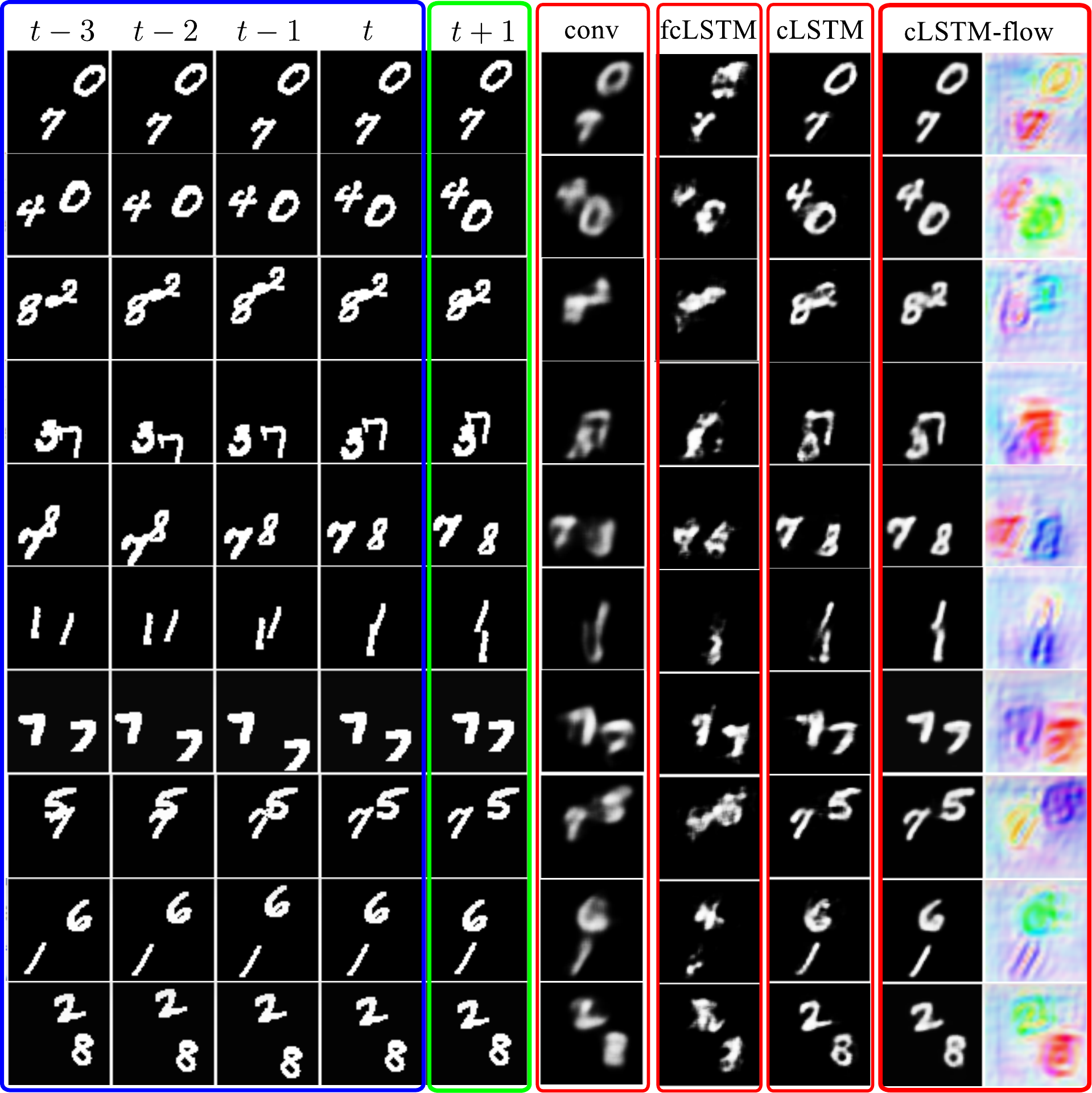}
\end{center}
   \caption{Qualitative results on moving MNIST test dataset. First four columns: the last frames (out of 10 in total) fed into the network (from $t-3$ to $t$); 5th column: the ground truth next frame ($t+1$); 6th column: AE-Conv; 7th column: AE-fcLSTM; 8th column: AE-ConvLSTM; last two columns: AE-ConvLSTM-flow predicted next frame and optical flow.}
\label{fig:prflow} 
\end{figure}

The loss function for this experiment was the binary cross-entropy \cite{SrivastavaMS15}, and the sequences are fed in as binary data. Table \ref{tab:all_params} presents the test errors and Figure \ref{fig:prflow} shows qualitative results. All the architectures using LSTM blocks report lower errors than AE-Conv, showing the importance of exploiting the temporal dependency between frames. Although the results of AE-Conv seem qualitatively better than AE-fcLSTM, they are very blurred, leading to the increased error. The architectures using the proposed convolutional LSTM units are more efficient than the classic fully-connected units, since the former preserve the spatial information; this can be observed in both the quantitative and the qualitative results. Importantly, AE-ConvLSTM-flow produces boundaries with better continuity than AE-ConvLSTM, suggesting that it acquires a basic form of objectness awareness, \ie it learns to reason about objects as a whole due to their motion. Equally, these results show that the optical flow maps produced by our network correspond to a basic form of image segmentation into objects identified by their movement. Hence, this setup could potentially be more useful for supervised video segmentation than a classic video autoencoder.

\begin{figure}[t]
\begin{center}
   \includegraphics[width=\linewidth]{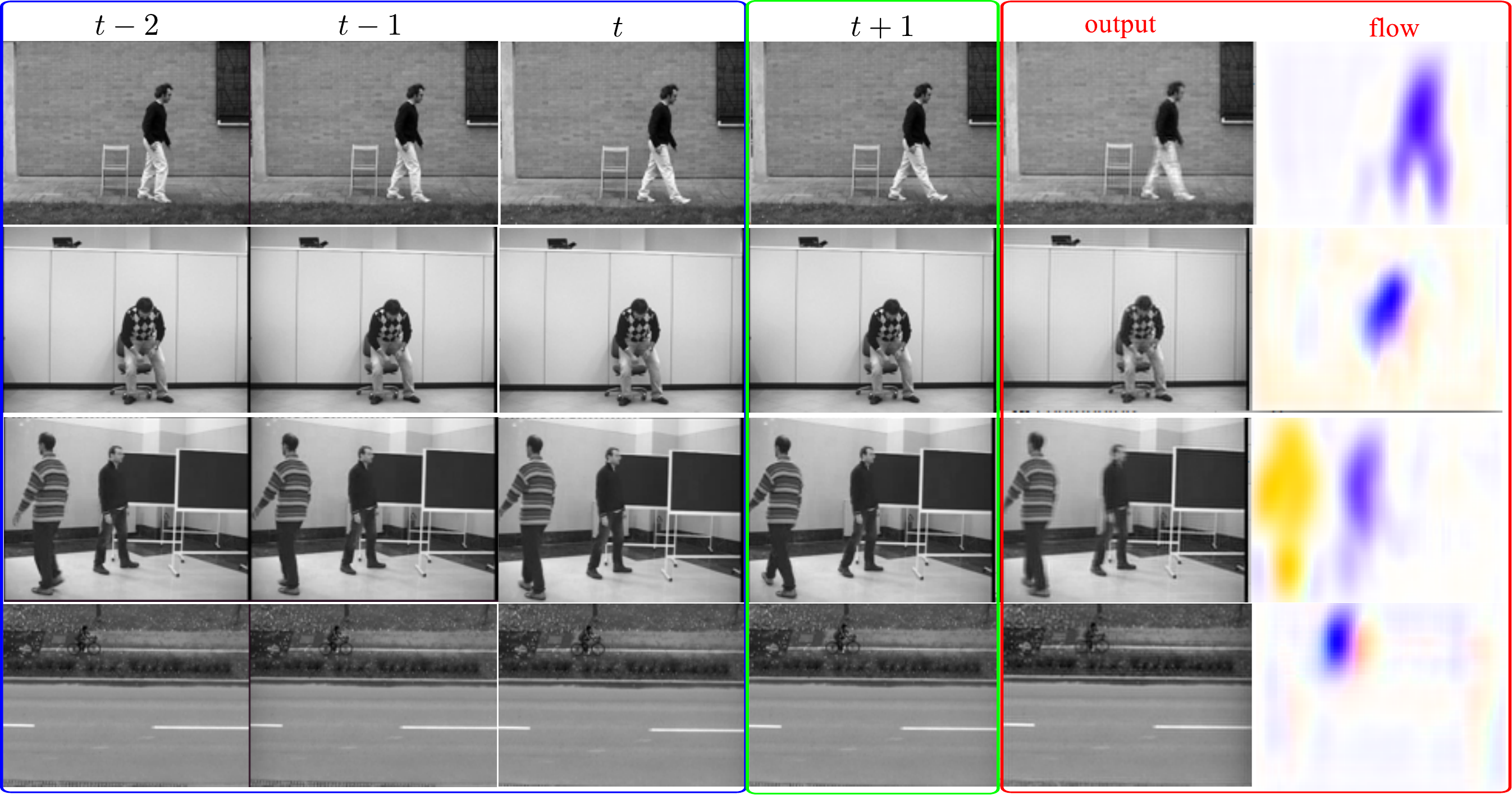}
\end{center}
   \caption{Qualitative results on real videos. The first three columns show the last frames (out of 12 in total) fed into the network (from $t-2$ to $t$), the $t+1$ column shows the ground truth next frame, and the last two columns show the predicted next frame and the predicted optical flow.}
\label{fig:prflowreal} 
\end{figure}

\paragraph{Real dataset.}
We train our architecture on real videos extracted from HMDB-51 dataset \cite{Kuehne11}, with about 107k frames in total (1.1 hours), and present qualitative results of the predicted next frame and the optical flow maps on test sequences extracted from PROST \cite{Santner2010} and ViSOR \cite{Vezzani2010} datasets. Figure \ref{fig:prflowreal} illustrates results obtained on different test sequences and videos showing the optical flow for entire test sequences are included in the supplemental material. The results indicate that the moving elements of the scene are identified by the network. However, the accuracy of the estimated flow is not comparable to supervised setups for optical flow estimation. But this is expected since in our case we predict the flow map that links the already seen frame $t$ to the unseen frame $t+1$, whereas classic optical flow methods estimate the optical flow between two already seen frames. Intuitively, we expect our network to extract more meaningful features than a classic supervised setup, since the problem is more challenging, at least for articulated objects. 


\subsection{Application to weakly-supervised video semantic segmentation}
\begin{figure}[h!]
\begin{center}
    \includegraphics[width=\textwidth]{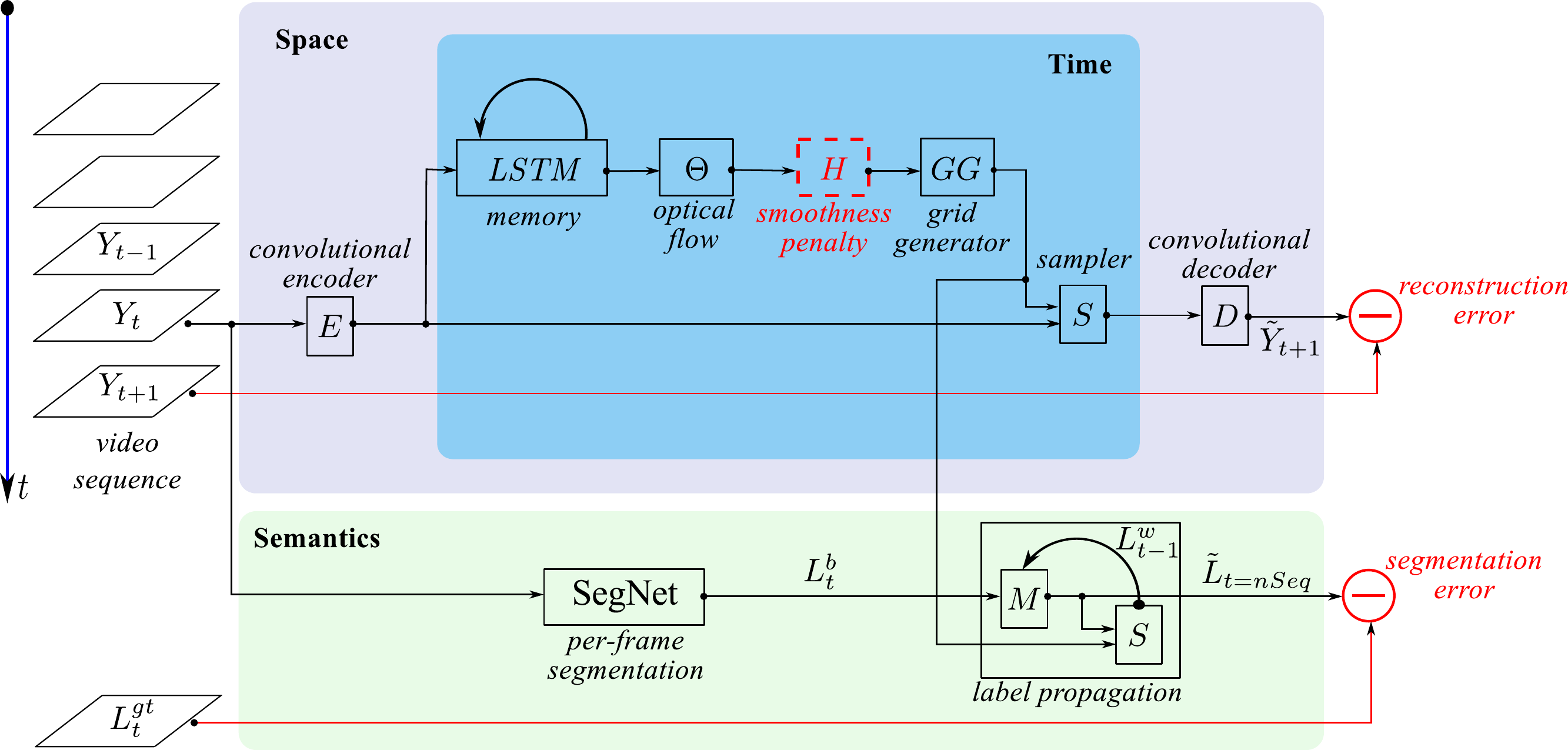}
    \caption{SegNet-flow: weakly-supervised video segmentation through label propagation using optical flow.}
    \label{fig:wsvs}
\end{center}%
\end{figure}

Almost no labelled dataset of real videos exists for semantic segmentation\footnote{SceneNet \cite{Handa15} is a recent synthetic dataset allowing to obtain synthetic labelled videos by rendering labelled synthetic 3D scenes. However, the models lack texture. Sun3D \cite{Xiao13} contains video sequences with labels obtained by label propagation. However, the quality of the labelling is quite poor.}, limiting the applicability of deep architectures for the task. There exist, however, datasets that contain video sequences with one labelled frame per sequence \cite{Brostow08,Silberman12}. We take advantage of our autoencoder to enable weakly-supervised video segmentation. The proposed framework, denoted SegNet-flow, is sketched in Figure \ref{fig:wsvs}. We use AE-ConvLSTM-flow together with a per-frame semantic segmentation module, SegNet\footnote{Although not state-of-the-art for image semantic segmentation, we chose to use SegNet in our experiment due to its reduced memory requirements. This is the only setup allowing us to train the entire network on one GPU with 12G memory.} \cite{Badrinarayanan15}. Using the flow maps produced by AE-ConvLSTM-flow, we warp (propagate) SegNet's predictions until reaching the time step for which ground truth labels are available. The label propagation is performed by a customised recurrent module, which integrates a merging block $M$ and a sampler $S$ similar to the one used in AE-ConvLSTM-flow. $M$ is a recurrent convolutional module, which receives as input SegNet's basic predictions $L_t^b$ at time $t$, and the warped predictions $L_{t-1}^w$ from time $t-1$ generated by $S$, and produces the final label predictions $\tilde L_t$ for time $t$. For the time steps where ground truth labels $L_t^{gt}$ are available, we minimise the cross-entropy segmentation error between $\tilde L_t$ and $L_t^{gt}$. This error term is added to the existing reconstruction error term. In this way, the two tasks, flow prediction and semantic segmentation, are learnt jointly with a minimum of supervision required.  
 
\begin{figure}[h!]
\begin{center}
    \includegraphics[width=\textwidth]{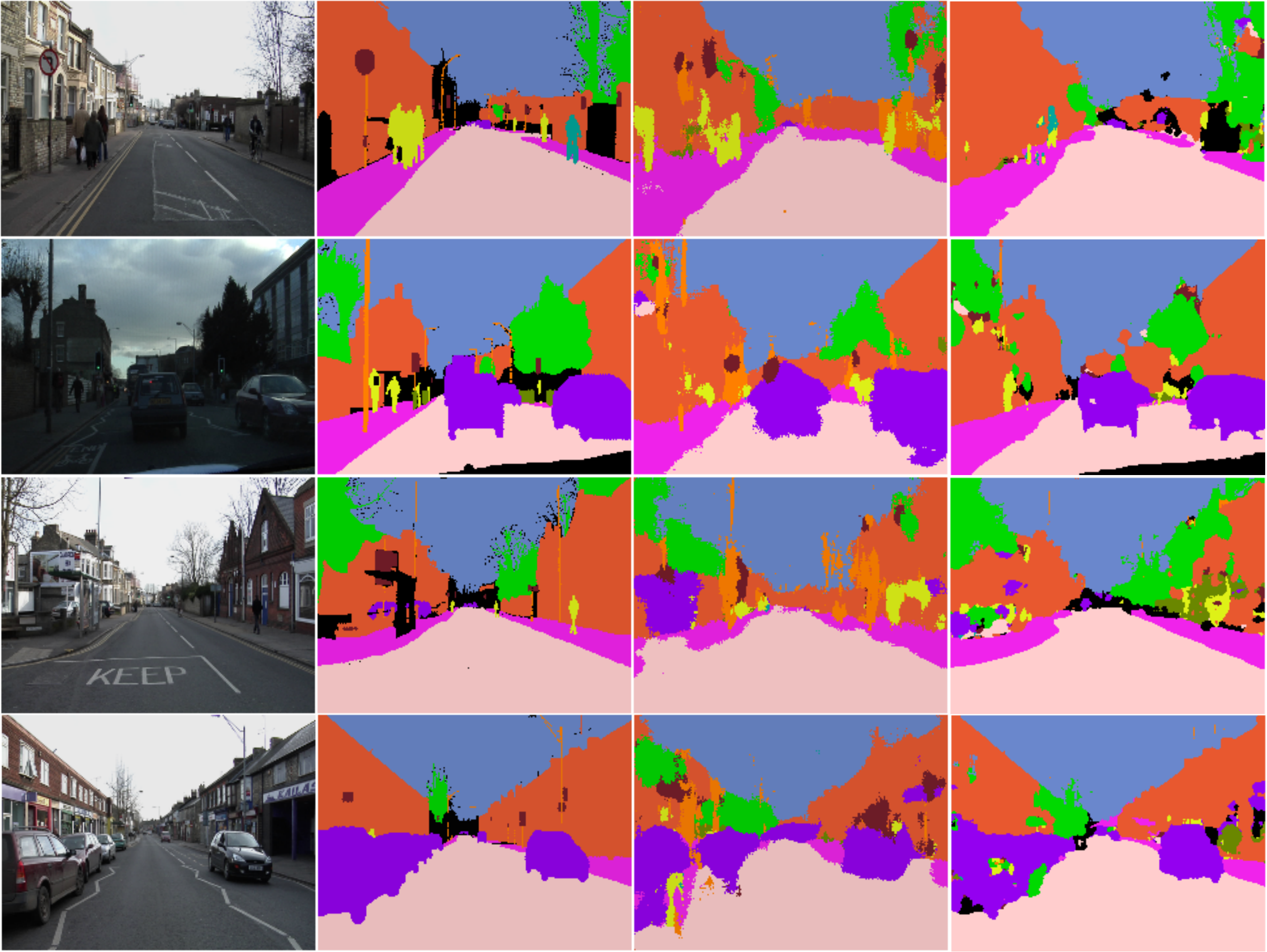}
    \caption{Qualitative results on Camvid. 1st column: input frame, 2nd column: ground truth, 3rd column: SegNet result, 4th column: SegNet-flow result. The flow constraints present in SegNet-flow make a significant difference in smoothing the segmentation results produced by SegNet. We expect this to hold irrespective of the segmentation module used.}
    \label{fig:segnet-flow-results}
\end{center}%
\end{figure}

We trained our architecture on Camvid dataset \cite{Brostow08}, which consists of road scene RGB videos, with about 10k training frames in total, among which 468 are labelled, and about 7k testing frames, among which 233 are labelled. However, due to memory limitations, we were not able to use all the (unlabelled) available frames, but only sequences of 6 frames around the labelled ones. Hence about 2.5k frames were used for training. Additionally, we downsampled the videos to $240\times 180$ pixels per frame\footnote{Note that the results in Table \ref{tab:segacc} for basic SegNet are different from those reported in \cite{Badrinarayanan15}, due to this downsampling.}. We use the 12 classes labelling, similar to \cite{Badrinarayanan15}. Table \ref{tab:segacc} summarises the results of basic SegNet on the test set compared to our setup, and Figure \ref{fig:segnet-flow-results} shows qualitative results. It can be observed that the quality of the segmentation increases considerably for the large classes, but diminishes for the small thin structures. We believe the main reason for this is the relatively small size of the training set. SegNet-flow has about one million parameters more than the basic SegNet, hence the former model is less constrained. Equally, the fact that the flow estimation is performed on a downsampled version of the input (after the spatial encoder) could lead to a poor flow estimation for small structures. Nevertheless, we were able to train end-to-end an architecture performing joint flow estimation and semantic segmentation from videos. Understanding better and improving this framework constitute an exciting direction of future work. 

\begin{table}
\tiny{
\centering
\begin{tabular}{|l|l|l|l|l|l|l|l|l|l|l|l|l|l|} 
\hline
\textbf{Network}  & \cellcolor{skyColor}\rotatebox{90}{Sky} & \cellcolor{buildingColor}\rotatebox{90}{Building}  & \cellcolor{poleColor}\rotatebox{90}{Pole} & \cellcolor{roadColor}\rotatebox{90}{Road} & \cellcolor{sidewalkColor}\rotatebox{90}{Sidewalk}  & \cellcolor{vegetationColor}\rotatebox{90}{Vegetation}  & \cellcolor{tsignColor}\rotatebox{90}{Traffic sign}   & \cellcolor{fenceColor}\rotatebox{90}{Fence} & \cellcolor{carColor}\rotatebox{90}{Car} & \cellcolor{pedestrianColor}\rotatebox{90}{Pedestrian}   & \cellcolor{cyclistColor}\rotatebox{90}{Cyclist}  & \cellcolor{voidColor}\rotatebox{90}{Class Avg.} &  \cellcolor{voidColor}\rotatebox{90}{Global Avg.} \\ \hline
SegNet & 89.2 & 65.4 & \textbf{22.6} & 92.0 & \textbf{70.7} & 44.8 & \textbf{14.9} & \textbf{5.38} & 60.5 & \textbf{25.6} & \textbf{2.73} & \textbf{44.9} & 75.3 \\ \hline
SegNet-flow & \textbf{92.9} & \textbf{73.4} & 3.22 & \textbf{97.2} & 62.9 & \textbf{52.3} & 2.37 &  2.35 & \textbf{63.3} & 14.9 & 1.70 & 42.4 & \textbf{76.9} \\ 
\hline
\end{tabular}}
\vspace{0.5mm} \vspace{0.5mm}
\caption{Per-class and global average segmentation accuracy.}
\label{tab:segacc}
\end{table}

\section{Conclusion and Future Work}
\label{sec:conclusion}
We proposed an original spatio-termporal video autoencoder based on an end-to-end differentiable architecture that allows unsupervised training for motion prediction. The core of the architecture is a module implementing a convolutional version of long short-term memory (LSTM) cells to be used as a form of artificial visual short-term memory. We showed that the proposed convolutional LSTM module performs better than existing classic autoencoders and than fully-connected LSTM implementations, while having a reduced number of parameters. The usefulness of the overall setup was illustrated in the task of weakly-supervised video semantic segmentation. 


We believe that our work can open up the path to a number of exciting directions. Due to the built-in feedback loop, various experiments can be carried out effortlessly, to develop further the basic memory module that we proposed. In particular, investigating on the size of the memory and its resolution/downsampling could shed some light into the causes of some geometric optical illusions; \eg when storing in memory an overly-downsampled version of the visual input, this could affect the capacity of correctly perceiving the visual stimuli, in accordance with Shannon-Nyquist sampling theorem \cite{Shannon49}. Equally, we hypothesise that our convolutional version of LSTM cells can lead to ambiguous memory activations, \ie the same activation would be produced when presenting a temporally moving boundary or a spatially repeated boundary. This could imitate the illusory motion experienced by biological VSTMs, \ie static repeated patterns that produce a false perception of movement \cite{Conway08062005}.  


A necessary future development is the integration of the memory module with an attention mechanism and a form of long-term memory module, to enable a complete memory system, indispensable for supervised tasks.  

Last but not least, the proposed architecture composed of memory module and built-in feedback loop could be applied for static images as a compression mechanism, similar to the inspirational work on jigsaw video compression \cite{Kannan2007}. The memory addressing in that case could use a content-based mechanism similar to NTM \cite{GravesWD14}.

\subsubsection*{Acknowledgments}
We are greatly indebted to the \emph{Torch} community for their efforts to maintain this great library, and especially to Nicholas L{\'e}onard for his helpful advice. We are grateful to CSIC--University of Cambridge for funding this work (grant number EP/L010917/1). 

\bibliography{egbib}
\bibliographystyle{iclr2016_workshop}

\end{document}